\documentclass[10pt,twocolumn,letterpaper]{article}

\usepackage[pagenumbers]{cvpr} 

\usepackage{float}
\usepackage{color}
\usepackage{graphicx}
\usepackage{amsmath}
\usepackage{booktabs}

\usepackage{colortbl}
\usepackage{booktabs}
\usepackage{multirow}
\usepackage[table,xcdraw]{xcolor}
\usepackage{stfloats}

\usepackage{algorithm}
\usepackage{algorithmic}

\usepackage{amsmath}

\usepackage{multirow}
\usepackage{graphicx}
\usepackage{makecell}
\usepackage{bbding}
\usepackage{xcolor}

\definecolor{cvprblue}{rgb}{0.21,0.49,0.74}
\usepackage[pagebackref,breaklinks,colorlinks,citecolor=cvprblue]{hyperref}

\hypersetup{
    colorlinks=true,            
    linkcolor=red,              
    citecolor=green            
}

\usepackage{cuted}
\usepackage{multirow}
\usepackage[accsupp]{axessibility} 

\usepackage{adjustbox}
\usepackage{soul}
\def\paperID{12278} 
\def\confName{CVPR}
\def\confYear{2025}

\title{2DGS-Room: Seed-Guided 2D Gaussian Splatting with Geometric Constrains for High-Fidelity Indoor Scene Reconstruction}

\author{
    Wanting Zhang\textsuperscript
    \quad Haodong Xiang\textsuperscript
    \quad Zhichao Liao\textsuperscript 
    \quad Xiansong Lai\textsuperscript
    \quad Xinghui Li\textsuperscript{\dag}
    \quad Long Zeng\textsuperscript{\dag}
    \\Tsinghua University \\
    \url{https://valentina-zhang.github.io/2DGS-Room/}
}

\begin{document}

\twocolumn[{%
    \maketitle
    \begin{figure}[H]
    \vspace{-0.8cm}
    \hsize=\textwidth 
    \centering
    \includegraphics[width=\textwidth]{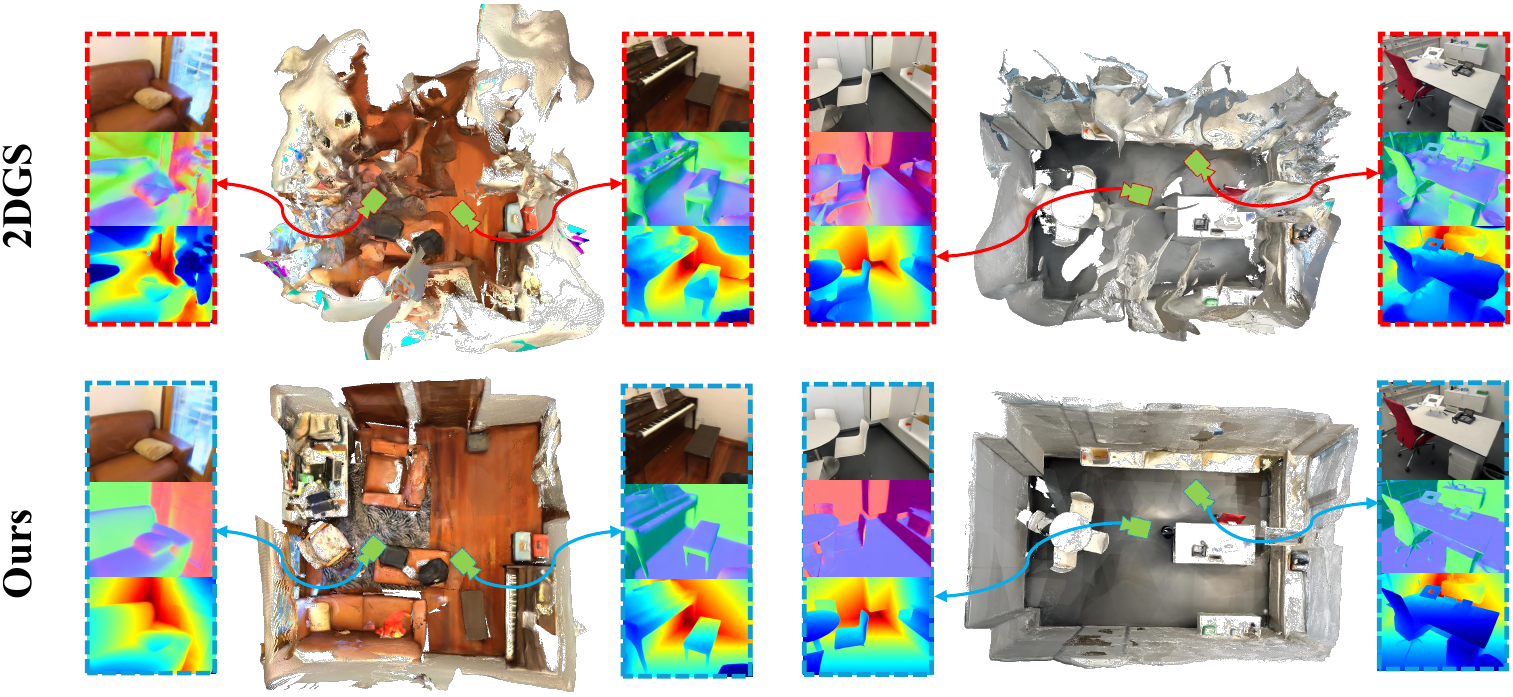}
    \vspace{-0.5cm}
    \caption{\textbf{2DGS-Room achieves high-fidelity geometric reconstructions for indoor scenes.} We introduce seed points to guide the distribution of 2D Gaussians coupled with geometric constraints, leading to clearer structures and more accurate geometry.}
    \label{fig:teaser}
    \end{figure}
}]

\begin{abstract}

The reconstruction of indoor scenes remains challenging due to the inherent complexity of spatial structures and the prevalence of textureless regions. Recent advancements in 3D Gaussian Splatting have improved novel view synthesis with accelerated processing but have yet to deliver comparable performance in surface reconstruction. In this paper, we introduce \textbf{2DGS-Room}, a novel method leveraging 2D Gaussian Splatting for high-fidelity indoor scene reconstruction. Specifically, we employ a seed-guided mechanism to control the distribution of 2D Gaussians, with the density of seed points dynamically optimized through adaptive growth and pruning mechanisms. To further improve geometric accuracy, we incorporate monocular depth and normal priors to provide constraints for details and textureless regions respectively. Additionally, multi-view consistency constraints are employed to mitigate artifacts and further enhance reconstruction quality. Extensive experiments on ScanNet and ScanNet++ datasets demonstrate that our method achieves state-of-the-art performance in indoor scene reconstruction.


\end{abstract}    
\section{Introduction}
\label{sec:intro}


3D reconstruction from multi-view RGB images is a fundamental task in the fields of computer vision and computer graphics. The reconstructed models can be utilized in a wide range of applications, including virtual reality, video games, autonomous driving, and robotics. Reconstructing indoor scenes is a challenging task in the field of 3D reconstruction, as indoor environments often contain large textureless regions. MVS-based methods \cite{schonberger2016pixelwise, luo2016efficient, yao2018mvsnet} often yield incomplete or geometrically flawed reconstructions, primarily due to the geometric ambiguities arising from the presence of textureless regions.

Recent advancements in neural-radiance-field-based methods \cite{wang2021neus, wang2022neuris, yu2022monosdf, li2023edge, li2024fine} that utilize signed distance fields (SDF) for scene modeling have enabled accurate and complete mesh reconstruction in indoor environments. This progress is attributed to the continuity of neural SDFs and the integration of monocular geometric priors \cite{yu2022monosdf}. Although neural-radiance-field-based methods achieve high-quality reconstruction, they are computationally expensive due to the need for dense ray sampling, resulting in long optimization times. Fortunately, 3D Gaussian Splatting (3DGS) \cite{kerbl20233dgs} enhances the optimization and rendering efficiency of neural rendering through its differentiable rasterization technique, offering new possibilities for 3D scene reconstruction. 2DGS \cite{huang20242d} build upon 3DGS by using 2D-oriented planar Gaussians as primitives, significantly improving surface reconstruction quality. Despite these advances, Gaussian splatting-based methods still often produce floating artifacts and incomplete reconstructions in indoor scenes, due to the lack of structured geometric constraints.


In this work, we present a novel approach named \textbf{2DGS-Room}, aiming to achieve high-fidelity geometric reconstruction for indoor scenes based on 2D Gaussian Splatting. Considering the scene’s underlying structure, we propose a seed-guided mechanism to control the distribution and density of 2D Gaussians. Specifically, we introduce a seed-guided initialization to generate 2D Gaussians, ensuring their alignment with scene surfaces to improve geometric accuracy. To further refine the reconstruction, we propose a seed-guided optimization strategy that dynamically adjusts seed point density through gradient-guided growth and contribution-based pruning, enabling efficient representation of fine details. Additionally, we incorporate monocular depth and normal priors to provide crucial geometric constraints. The depth prior addresses distortions in detailed areas, while the normal prior ensures accurate surface estimation in textureless regions. Furthermore, we introduce multi-view consistency constraints to address residual artifacts, which enforces both geometric and photometric consistency across multiple views. 


Extensive qualitative and quantitative experiments show that compared with Gaussian-based methods, 2DGS-Room achieves start-of-the-art performance in indoor scenarios. In summary, our contributions are as follows:

\begin{itemize}[leftmargin=2em]
  \item We propose \textbf{2DGS-Room}, a novel method for indoor scene reconstruction based on 2DGS, which leverages the seed points maintaining the scene structure to guide the distribution and density of 2D Gaussians.
  
  \item We introduce monocular depth and normal priors to provide geometric cues, improving the reconstruction of detailed areas and textureless regions respectively.

  \item We employ multi-view constraints incorporating geometric and photometric consistency to further enhance the reconstruction quality.
  
  \item Our method achieves high-quality surface reconstruction for indoor scenes. Extensive experiments on indoor scene datasets show that our method achieves state-of-the-art in multiple evaluation metrics.
\end{itemize}

\begin{figure*}[t]
  \centering
   \includegraphics[width=\linewidth]{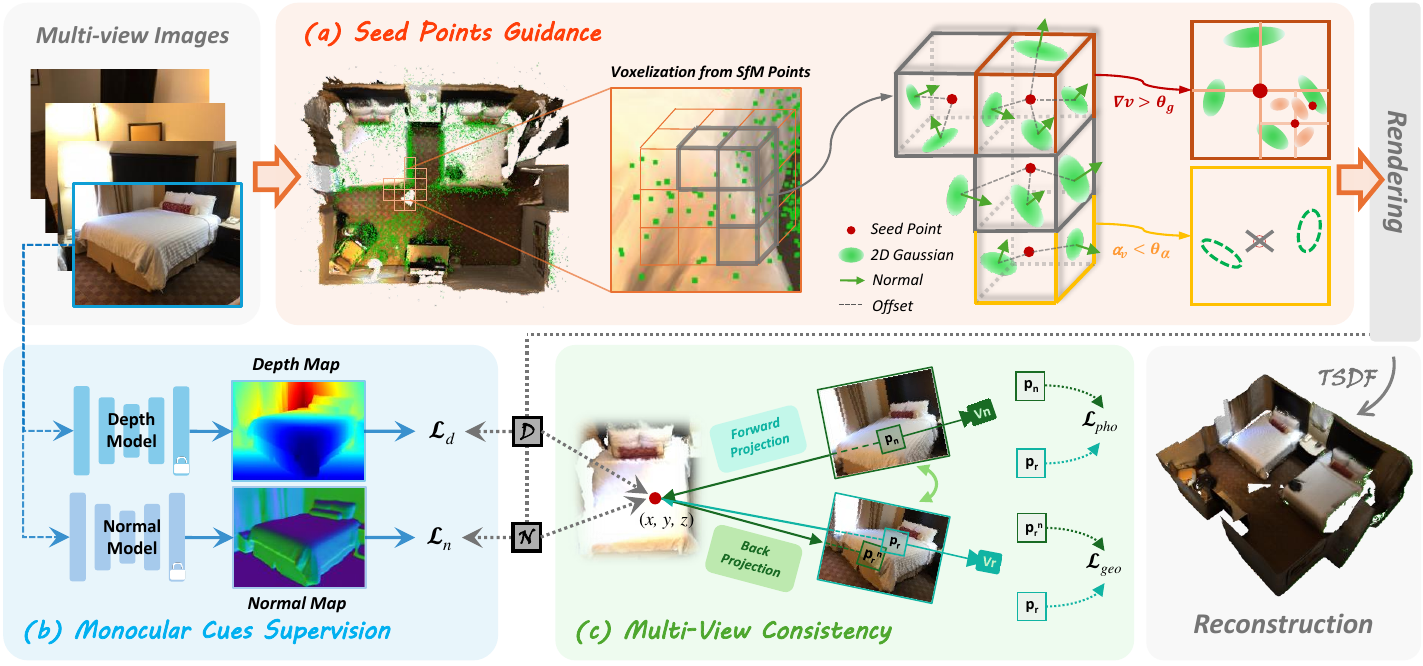}
   \caption{\textbf{Overview of 2DGS-Room.} Given multi-view posed images, we improve 2DGS to achieve high-fidelity geometric reconstruction for indoor scenes. (a) Starting from an SfM-derived point cloud, we generate a set of seed points through voxelization, establishing a stable foundation for guiding the distribution and density of 2D Gaussians. We further introduce an adaptive growth and pruning strategy to optimize seed points. (b) We incorporate depth and normal priors, addressing the challenges of detailed areas and textureless regions. (c) We introduce multi-view consistency constraints to further enhance the quality of the indoor scene reconstruction.}
   \label{fig:overview}
\end{figure*}

\section{Related work}
\label{sec:related}



\subsection{Multi-View Stereo}
Multi-view stereo (MVS) methods \cite{barnes2009patchmatch, galliani2016gipuma, schonberger2016pixelwise, stereopsis2010accurate} estimate the 3D coordinates of pixels and explicitly reconstruct objects and scenes by matching features across a collection of posed images. The surface is then obtained through the application of Poisson surface reconstruction \cite{kazhdan2013screened}. In indoor scenes, particularly in large texture-less regions, these methods frequently encounter difficulties due to the scarcity of features. Voxel-based approaches \cite{broadhurst2001probabilistic, de1999poxels, seitz1999photorealistic, liu2020dist} optimize spatial occupancy and color within a voxel grid, thus avoiding the challenges of feature matching. However, high-resolution memory constraints degrade reconstruction quality. Learning-based multi-view stereo methods \cite{luo2016efficient, ummenhofer2017demon, zagoruyko2015learning, riegler2017octnetfusion, huang2018deepmvs, yao2018mvsnet, yao2019recurrent, yu2020fast, zhang2020visibility} implicitly match corresponding multi-view features through neural networks, enabling end-to-end 3D reconstruction. Nonetheless, even with extensive training data, errors may still occur in the results when handling occlusions, complex lighting, or regions with subtle textures.

\subsection{{Neural Radiance Field}}

Neural Radiance Fields (NeRF) \cite{mildenhall2021nerf} employs a multi-layer perceptron (MLP) to model a continuous volumetric function of density and color, enabling novel view synthesis through volume rendering. Methods such as Mip-NeRF \cite{barron2021mip, xu2022point, barron2023zip} enhance rendering quality by improving the ray sampling strategy. Other works \cite{muller2022instant, liu2020neural, fridovich2022plenoxels, chen2022tensorf, li2023neuralangelo} accelerate training and rendering through techniques such as multi-resolution hash encoding or resizing MLPs. Some studies aim to enhance rendering quality by incorporating regularization terms. For example, depth regularization \cite{deng2022depth, wei2021nerfingmvs} explicitly supervises ray termination to minimize unnecessary sampling time. Other approaches focus on enforcing smoothness constraints on rendered depth maps \cite{niemeyer2022regnerf} or utilizing multi-view consistency regularization in sparse-view scenarios \cite{wang2023sparsenerf, lao2024corresnerf}. Some research explores the use of alternative implicit functions to enhance the geometric reconstruction capabilities of NeRF, such as occupancy grids \cite{niemeyer2020differentiable, oechsle2021unisurf} and signed distance functions (SDFs) \cite{yariv2020multiview, wang2022neuris, wang2021neus, yariv2021volume, li2023neuralangelo}, replacing NeRF's volumetric density field. To further enhance reconstruction quality, \cite{fu2022geo, zhang2022critical} suggest regularizing optimization with SfM points, while \cite{guo2022neural, yu2022monosdf} incorporate priors like the Manhattan world assumption and pseudo depth supervision. However, these approaches often lead to incomplete reconstructions and require extensive optimization time.

\subsection{{Gaussian Splatting}}

3D Gaussian Splatting \cite{kerbl20233dgs} explicitly represents 3D scenes using learnable Gaussian primitives, enabling high-quality novel view synthesis with short training times and high rendering frame rates. The 3DGS method is solely responsible for the image loss, and after initializing with sparse point clouds generated by SfM \cite{schonberger2016structure}, no further constraints are applied to the Gaussian primitives. This leads to a disorganized distribution of the optimized Gaussian primitives, resulting in poor geometric properties. Works such as DN-Splatter \cite{turkulainen2024dn}, GaussianRoom \cite{xiang2024gaussianroom} and GSDF \cite{yu2024gsdf} introduce geometric priors or leverage the accurate geometric information from SDFs to supervise the optimization of Gaussians. SuGaR \cite{guedon2023sugar}, PGSR \cite{chen2024pgsr} and RaDe-GS \cite{zhang2024rade} use Flatten Gaussians to represent scenes, enhancing surface reconstruction capabilities. In contrast, 2DGS \cite{huang20242d} directly applies 2D oriented planar Gaussians instead of 3D Gaussian primitives to represent 3D scenes, achieving better surface reconstruction results. 
However, it still encounters poor reconstruction in indoor scenes due to Gaussian primitives lacking geometric constraints. 



\section{Preliminary}

The key innovation of 2DGS \cite{huang20242d} lies in its transformation of 3D volumetric Gaussians into flat 2D Gaussians, or surfels, for scene representation. It directly models scenes with 2D elliptical disks, simplifying the representation process and yielding more accurate geometry without extra mesh refinement.

Each 2D Gaussian disk, defined in a local tangent plane, is parameterized by a central point $ \mathbf{p}_k $, two orthogonal tangential vectors $ \mathbf{t}_u $ and $ \mathbf{t}_v $, and a scaling vector $ (s_u, s_v) $ that controls the variances along each direction. The normal $ \mathbf{t}_w $ of each Gaussian disk is computed as $ \mathbf{t}_w = \mathbf{t}_u \times \mathbf{t}_v $ and this orientation can be arranged into a rotation matrix $ \mathbf{R} = [\mathbf{t}_u, \mathbf{t}_v, \mathbf{t}_w] $. The scaling factors can be arranged into a 3 × 3 diagonal matrix $ \mathbf{S} = [s_u, s_v, 0] $. Then a 2D Gaussian can be parameterized:
\begin{equation}
    P(u,v) =\mathbf{p}_{k}+s_{u}\mathbf{t}_{u}u+s_{v}\mathbf{t}_{v}v=\mathbf{H}(u,v,1,1),
\end{equation}
where $\mathbf{H} \in 4 \times 4$ is a homogeneous transformation matrix representing the geometry of the 2D Gaussian:
\begin{equation}
    \mathbf{H}=\begin{bmatrix}s_{u} \mathbf{t}_u &s_{v} \mathbf{t}_v &{\mathbf{0}}& \mathbf{p}_k\\{0}&{0}&{0}&{1}\end{bmatrix}=\begin{bmatrix}{\mathbf{RS}}& \mathbf{p}_k\\{\mathbf{0}}&{1}\end{bmatrix}.
\end{equation}

In the Gaussian’s tangent frame $ (u, v) $, the 2D Gaussian value $ \mathcal{G}(\mathbf{u}) $ at point $ \mathbf{u} = (u, v) $ is evaluated as:
\begin{equation}
    \mathcal{G}(\mathbf{u}) = \exp\left(-\frac{u^2 + v^2}{2}\right).
\end{equation}


For efficient rendering, each 2D Gaussian is projected onto the image plane by a general 2D-to-2D mapping in homogeneous coordinates. Given a world-to-screen transformation matrix $ \mathbf{W} $, the screen space points can be derived from:
\begin{equation}
    \mathbf{x} = (xy, yz, z, z)^{\top} = \mathbf{WH} (u, v, 1, 1)^{\top}.
\end{equation}

\noindent where $\mathbf{x}$ represents a homogeneous ray emitted from the camera and passing through pixel $ (x, y) $ and intersecting the splat at depth $z$.

To avoid numerical instability, a ray-splat intersection is calculated explicitly by finding the intersection of three non-parallel planes in the 3D scene. Given an image coordinate $ \mathbf{x} = (x, y) $, the ray of a pixel can be defined by the intersection of two homogeneous planes: the x-plane $ \mathbf{h}_{x} = (-1, 0, 0, x) $ and the y-plane $ \mathbf{h}_{y} = (0, -1, 0, y) $. To compute the intersection with the Gaussian splat, both planes are transformed to $ uv $-space:
\begin{equation}
    \mathbf{h}_u = (\mathrm{W} \mathrm{H})^{\top} \mathrm{h}_x, \quad \mathbf{h}_v = (\mathrm{W} \mathrm{H})^{\top} \mathrm{h}_y.
\end{equation}

By homography, the two planes are used to find the intersection point $ (u(x), v(x)) $ with the 2D Gaussian splats, given by:
\begin{equation}
    u(\mathbf{x})=\frac{\mathbf{h}_{u}^{2}\mathbf{h}_{v}^{4}-\mathbf{h}_{u}^{4}\mathbf{h}_{v}^{2}}{\mathbf{h}_{u}^{1}\mathbf{h}_{v}^{2}-\mathbf{h}_{u}^{2}\mathbf{h}_{v}^{1}}, \quad v(\mathbf{x})=\frac{\mathbf{h}_{u}^{4}\mathbf{h}_{v}^{1}-\mathbf{h}_{u}^{1}\mathbf{h}_{v}^{4}}{\mathbf{h}_{u}^{1}\mathbf{h}_{v}^{2}-\mathbf{h}_{u}^{2}\mathbf{h}_{v}^{1}},
\end{equation}
\noindent where $ \mathbf{h}_{u}^{i} $ and $ \mathbf{h}_{v}^{i} $ are components of the transformed planes in the Gaussian’s tangent frame.

\section{Methods}


Given multi-view posed images, our goal is to optimize 2DGS \cite{huang20242d} to accurately reconstruct the geometry of indoor scenes. To this end, we first propose a seed-guided mechanism, which leverages seed points to control the distribution and density of 2D Gaussians, thereby improving the accuracy and efficiency of scene representation in indoor scenes (Sec. \ref{subsec:seed}). To further improve geometric accuracy, we incorporate depth and normal priors, which enhance the representation of detailed areas and textureless regions, respectively (Sec. \ref{subsec:prior}). Finally, to mitigate floating artifacts caused by lighting variations in indoor scenes, we introduce multi-view consistency constraints, further enhancing the quality of the indoor scene reconstruction (Sec. \ref{subsec:mv}). An overview of our framework is provided in Fig. \ref{fig:overview}.

\subsection{Seed Points Guidance}
\label{subsec:seed}



\begin{figure}[t]
  \centering
   \includegraphics[width=\linewidth]{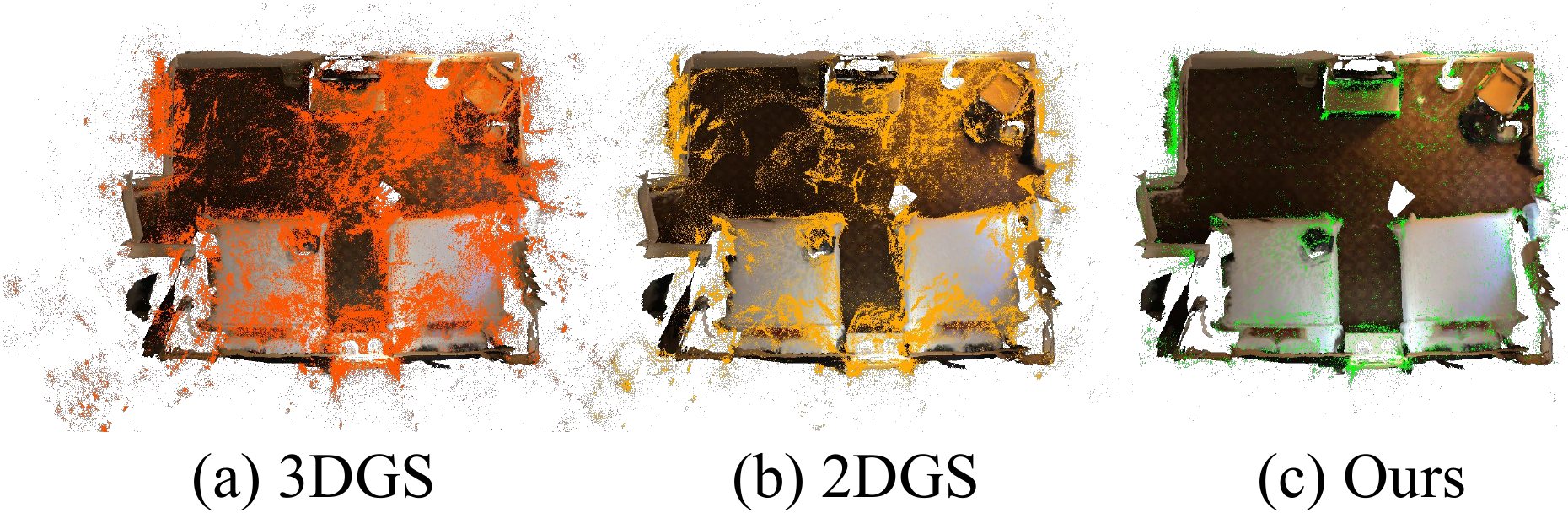}
   \caption{\textbf{Ground truth scene surface and Gaussian primitives distribution.} Compared with 3DGS and 2DGS, our method significantly reduces scattered floaters in the non-surface areas, benefitting from our designed structured geometric constraints. }
   \vspace{-0.3cm}
   \label{fig:seed}
\end{figure}

Existing methods \cite{kerbl20233dgs, huang20242d} tend to optimize Gaussians relying on each training view, ignoring the underlying structure of the scene. As illustrated in Fig.~\ref{fig:seed} (a) and (b), the Gaussian primitives fail to align with the surfaces. To overcome this limitation, we propose a seed-guided mechanism to control the distribution of 2D Gaussians. Specifically, we utilize a set of seed points to provide a stable foundation for generating 2D Gaussians, ensuring that the reconstruction reflects the underlying scene structure more accurately. Additionally, we introduce an adaptive growth and pruning strategy to dynamically adjust the density of seed points.



\noindent \textbf{Seed-Guided Initialization.} Starting from an SfM-derived point cloud $ \mathbf{P} \in \mathbb{R}^{M \times 3} $, we first filter some unreliable outliers. We define a confidence measure $ O_{\mathbf{p}_i} $ for each individual point $ \mathbf{p}_i $ in the point cloud. This measure is expressed as follows:
\begin{equation}
    O_{\mathbf{p}_i} = 
    \begin{cases}
    1 & \text{if } m \geq \epsilon \\
    0 & \text{if } m < \epsilon 
    \end{cases},
\end{equation}
where $ m $ represents the number of image feature matches associated with $ \mathbf{p}_i $, and $ \epsilon $ is a predefined threshold. Points with a number of matched features below $ \epsilon $ are deemed unreliable and removed from the point cloud to ensure a more accurate reconstruction.

Following the filtering process, we apply voxelization to generate a set of seed points $ \mathbf{V} \in \mathbb{R}^{N \times 3} $ by selecting the center points of each voxel grid to represent the seed points:
\begin{equation}
\mathbf{V} = \left\{ \left\lfloor \frac{\mathbf{P}}{\delta} \right\rfloor \cdot \delta \right\},
\end{equation}

\noindent where $ \delta $ denotes the voxel grid size. Each seed point $ v \in \mathbf{V} $ serves as the basis for deriving several 2D Gaussians, which are positioned based on learnable offsets from the seed point. This initialization ensures that the distribution of Gaussians is closely aligned with the underlying geometry of the scene, thereby improving the overall robustness of the reconstruction quality.

For each seed point $ v \in \mathbf{V} $, we initialize a set of $ k $ 2D Gaussians $ \{ \mathcal{G}_{i,j} \} $, where $ \mathcal{G}_{i,j} $ denotes the $ j $-th Gaussian associated with the $ i $-th seed. The position of each Gaussian is determined by a learnable offset $ \mathbf{O}_{i,j} $ from the seed point location:
\begin{equation}
    \mathbf{p}_{i,j} = \mathbf{v}_i + \mathbf{O}_{i,j},
\end{equation}

\noindent where $ \mathbf{p}_{i,j} \in \mathbb{R}^3 $ represents the global position of the Gaussian, and $ \mathbf{O}_{i,j} \in \mathbb{R}^3 $ is a learnable offset which is optimized during training to adjust each Gaussian’s local position for better alignment with the scene.

Expect for the center position, each 2D Gaussian is parameterized by the scaling $ \mathbf{s} \in \mathbb{R}^2 $, rotation $ \mathbf{t} \in \mathbb{R}^2 $, appearance $ \mathbf{c} \in \mathbb{R}^3 $ and opacity $ \alpha \in \mathbb{R} $. At initialization, the scaling and rotation are aligned with the local geometry derived from the point cloud, which provides a starting approximation that reflects the scene’s spatial distribution. During training, these parameters are iteratively optimized to refine the representation.


\noindent \textbf{Seed-Guided Optimization.} In order to capture different levels of detail in complex indoor scenes, we develop an adaptive approach to dynamically adjust seed point density by combining gradient-guided growth and contribution-based pruning.

We utilize a gradient-guided growth strategy to increase seed point density adaptively, especially in areas with high structural complexity or fine details. For each voxel, we compute the average gradient $ \nabla v $ of the included 2D Gaussians across $N_g$ training iterations, using it as an indicator of structural complexity. When $ \nabla v $ exceeds a threshold $ \theta_g $, additional seed points are introduced to enhance representation. This growth occurs within a multi-resolution voxel structure, with thresholds that adapt according to the resolution level, ensuring a higher seed density in regions requiring more detail.

Moreover, we implement a contribution-based pruning strategy that selectively removes low-impact seed points. For each seed, we calculate the cumulative opacity $ \alpha_v $ of the connected 2D Gaussians over $N_\alpha$ iterations. If $ \alpha_v $ is below a predefined threshold $ \theta_{\alpha} $, the seed point is pruned, as its minimal contribution to scene opacity suggests the limited impact on the overall representation. This strategy allows us to allocate Gaussians to regions of higher structural significance, enhancing both computational efficiency and reconstruction quality.

\subsection{Monocular Cues Supervision}
\label{subsec:prior}

While the control of seed points enhances the structural consistency of the scene, it remains insufficient for achieving highly accurate geometry, particularly in detailed or textureless regions which are common in indoor environments. Therefore, we incorporate depth and surface normal priors, providing geometric constraints to further improve the scene reconstruction.

\noindent \textbf{Monocular Depth Supervision.} The depth prior is leveraged to mainly refine the spatial alignment of objects in the scene by aligning the rendered depths with reference depths predicted from a pre-trained model \cite{depth-pro}. We incorporate depth supervision by aligning the rendered depths with reference depths through a scale-and-shift-invariant loss \cite{depthloss}, compensating for relative scaling discrepancies that may arise in the representation of complex indoor geometries. 

Given the rendered depths $ \mathcal{\hat{D}} $, we first compute optimal scale $ s $ and shift $ t $ values to minimize discrepancies in scale and translation between our rendered depths and the reference depths to address potential inconsistencies that may arise due to relative scaling differences in complex scenes. Then we adjust the predicted depth map to obtain the aligned prediction: $ \mathcal{\hat{D}}_{\text{aligned}} = s \cdot \mathcal{\hat{D}} + t $.

The depth loss $ \mathcal{L}_{d} $ consists of two terms: a data term that minimizes the mean squared error (MSE) between the aligned rendered depths $ \mathcal{\hat{D}}_{\text{aligned}}$ and the reference depths $ \mathcal{D} $, and a regularization term for gradient consistency that encourages local smoothness in the depth rendering. Formally, the depth loss is defined as:
\begin{equation}
\mathcal{L}_{d} = \frac{1}{|\mathcal{V}_d|} \sum \|\mathcal{\hat{D}}_{\text{aligned}} - \mathcal{D}\|^2 + \lambda_{grad} \cdot \mathcal{L}_{\text{grad}},
\end{equation}

\noindent where $ |\mathcal{V}_d| $ represents the number of pixels with valid depths, and $ \mathcal{L}_{\text{grad}} $ is a spatial regularization term that penalizes abrupt depth variations across neighboring pixels.

\noindent \textbf{Monocular Normal Supervision.} Additionally, the normal prior plays a crucial role in addressing the reconstruction challenges of textureless or planar regions like walls and floors. So we also incorporate normal supervision to enforce a smooth and realistic surface orientation throughout the scene.

Let $ \mathcal{\hat{N}} $ denote the reference normals derived from a pre-trained model \cite{sne}, and $ \mathcal{N} $ represents the rendered normals. We first use the $\mathcal{L}_1$ norm loss to quantify the absolute difference in magnitude between the rendered and reference normals, promoting consistency in the length of the vectors:
\begin{equation}
\mathcal{L}_{1} = \frac{1}{|\mathcal{V}_n|} \sum \left| \mathcal{N} - \mathcal{\hat{N}} \right|,
\end{equation}
where $ |\mathcal{V}_n| $ is the number of pixels with valid reference normals.

To further encourage the alignment of $ \mathcal{} $ with $ \mathcal{\hat{N}} $, we use a cosine similarity loss that penalizes angular differences between the two normal vectors:
\begin{equation}
\mathcal{L}_{\cos} = \frac{1}{|\mathcal{V}_n|} \sum \left(1 - \frac{\mathcal{N} \cdot \mathcal{\hat{N}}}{\|\mathcal{N}\| \cdot \|\mathcal{\hat{N}}\|}\right).
\end{equation}

The final normal supervision loss $ \mathcal{L}_{n} $ is defined as:
\begin{equation}
\mathcal{L}_{n} =  \lambda_{1} \cdot \mathcal{L}_1 + \lambda_{\cos} \cdot \mathcal{L}_{\cos}.
\end{equation}

\setlength{\tabcolsep}{1.4mm}{
\begin{table*}[bp]
    \centering
    \renewcommand{\arraystretch}{1.0}
    {
    \begin{tabular}{l|ccccc|ccccc}
        \toprule
                                 & \multicolumn{5}{c|}{ScanNet \cite{scannet}}                                                                                                                                  & \multicolumn{5}{c}{ScanNet++ \cite{scannet++}}                                                                                                                                 \\
        \multirow{-2}{*}{Method} & Acc. ↓                     & Comp. ↓                   & Prec. ↑                    & Recall ↑                       & F-score ↑                      & Acc. ↓                     & Comp. ↓                   & Prec. ↑                    & Recall ↑                       & F-score ↑                      \\ 
        \midrule
        NeuS \cite{wang2021neus}                     & \cellcolor[HTML]{FFD8B2}0.105 & \cellcolor[HTML]{FFFFB2}0.124 & \cellcolor[HTML]{FFD8B2}0.448 & \cellcolor[HTML]{FFFFB2}0.378 & \cellcolor[HTML]{FFFFB2}0.409 & \cellcolor[HTML]{FFFFB2}0.160 & 0.224                         & 0.294                         & \cellcolor[HTML]{FFFFB2}0.221 & \cellcolor[HTML]{FFFFB2}0.251 \\
        Neuralangelo \cite{li2023neuralangelo}             & 0.185                         & 0.223                         & 0.252                         & 0.260                         & 0.255                         & 0.363                         & 0.264                         & 0.172                         & 0.120                         & 0.141                         \\
        3DGS \cite{kerbl20233dgs}                     & 0.338                         & 0.406                         & 0.129                         & 0.067                         & 0.085                         & \cellcolor[HTML]{FFB2B2}0.144 & 0.990                         & 0.322                         & 0.066                         & 0.104                         \\
        SuGaR \cite{guedon2023sugar}                    & 0.167                         & 0.148                         & 0.361                         & 0.373                         & 0.366                         & \cellcolor[HTML]{FFD8B2}0.158 & \cellcolor[HTML]{FFD8B2}0.178 & \cellcolor[HTML]{FFD8B2}0.383 & \cellcolor[HTML]{FFD8B2}0.349 & \cellcolor[HTML]{FFD8B2}0.361 \\
        2DGS \cite{huang20242d}                     & 0.157                         & 0.151                         & 0.336                         & 0.347                         & 0.341                         & 0.359                         & 0.228                         & 0.230                         & 0.160                         & 0.183                         \\
        PGSR \cite{chen2024pgsr}                     & \cellcolor[HTML]{FFFFB2}0.125 & \cellcolor[HTML]{FFD8B2}0.117 & \cellcolor[HTML]{FFFFB2}0.420 & \cellcolor[HTML]{FFD8B2}0.433 & \cellcolor[HTML]{FFD8B2}0.426 & 0.204                         & \cellcolor[HTML]{FFFFB2}0.202 & \cellcolor[HTML]{FFFFB2}0.353 & 0.217                         & 0.249                         \\
        RaDe-GS \cite{zhang2024rade}                  & 0.167                         & 0.205                         & 0.309                         & 0.307                         & 0.306                         & 0.284                         & 0.252                         & 0.171                         & 0.179                         & 0.166                         \\
        \textbf{2DGS-Room (Ours)}                     & \cellcolor[HTML]{FFB2B2}0.055 & \cellcolor[HTML]{FFB2B2}0.092 & \cellcolor[HTML]{FFB2B2}0.648 & \cellcolor[HTML]{FFB2B2}0.518 & \cellcolor[HTML]{FFB2B2}0.575 & 0.262                         & \cellcolor[HTML]{FFB2B2}0.112 & \cellcolor[HTML]{FFB2B2}0.450 & \cellcolor[HTML]{FFB2B2}0.498 & \cellcolor[HTML]{FFB2B2}0.464 \\ 
        \bottomrule
    \end{tabular}
    } 

    \caption{\textbf{Quantitative reconstruction comparison on ScanNet and ScanNet++ dataset.} Averaged results are reported over 8 scenes and 4 scenes, respectively. 2DGS-Room achieves the best F-score.}
    \label{tab:2D_compare_quality}
\end{table*}
}

\subsection{Multi-View Consistency Constraints}
\label{subsec:mv}

The strategies outlined above significantly improve the accuracy of indoor scene reconstruction, but we observe that some small floaters may still persist in certain scenarios. These cases are likely caused by the complex lighting variations and subtle spatial structures typical in indoor environments. Therefore, we introduce multi-view consistency constraints to further refine the reconstruction by reducing the inconsistencies that occasionally manifest across different views. Specifically, as shown in Figure \ref{fig:overview}, given a reference view $ V_r $, we select a neighboring view $V_n$ and enforce geometric consistency and photometric consistency between the two views.


\noindent \textbf{Geometric Consistency Constraint.} To ensure consistent geometry across views, we define a pixel-wise geometric consistency loss that penalizes discrepancies in the forward and backward projections for each individual pixel.

We compute a transformation $ H_{rn} $ to represent the homography matrix mapping a pixel $ \mathbf{p}_r $ from $ V_r $ to the corresponding pixel $ \mathbf{p}_n $ in $ V_n $:
\begin{equation}
H_{rn} = K_n \left( R_{rn} - \frac{T_{rn} \mathcal{N}_r^\top}{\mathcal{D}_r} \right) K_r^{-1},
\end{equation}

\noindent where $ K $ denotes the camera's intrinsic matrix. $ R_{rn} $ and $ T_{rn} $ are the relative rotation and translation from the reference frame to the neighboring frame.

For each pixel $ \mathbf{p}_r $ , we project it forward from $ V_r $ to $ V_n $ using $ H_{rn} $, and then back-project from $ V_n $ to $ V_r $ using $ H_{nr} $. The resulting multi-view geometric consistency loss $ \mathcal{L}_{\text{geo}} $ is formulated as:
\begin{equation}
\mathcal{L}_{geo} = \frac{1}{|\mathcal{V}_e|} \sum_{ \mathbf{p}_r \in \mathcal{V}_e } \| \mathbf{p}_r - H_{nr} H_{rn} \mathbf{p}_r \|,
\end{equation}

\noindent where $ \mathcal{V}_e $ is a set of valid pixels excluding those with high forward and backward projection errors.

\noindent \textbf{Photometric Consistency Constraint.} To account for local variations in texture and illumination, we also enforce photometric consistency which is measured using the normalized cross-correlation (NCC) \cite{ncc}, penalizing differences in pixel intensity distributions between the views. 

Focusing on geometric details, we convert color images into grayscale and the photometric consistency loss $\mathcal{L}_{pho}$ is defined as:
\begin{equation}
\mathcal{L}_{pho} = \frac{1}{|\mathcal{V}_e|} \sum_{ \mathbf{p}_r \in \mathcal{V}_e } \left( 1 - \text{NCC}(G_r(\mathbf{p}_r), G_n(H_{rn}\mathbf{p}_r)) \right),
\end{equation}

\noindent where $ G_r $ and $ G_n $ denote the grayscale intensities of the patches in $ V_r $ and $ V_n $, respectively.


Finally, the total multi-view consistency loss $ \mathcal{L}_{mv} $ is given by:
\begin{equation}
\mathcal{L}_{mv} = \lambda_{geo} \mathcal{L}_{geo} + \lambda_{pho} \mathcal{L}_{pho}.
\end{equation}

\subsection{Optimization}

In summary, with $ \mathcal{L}_{rgb} $ representing the photometric supervision that minimizes the difference between rendered and input images proposed in the original 2DGS, our final training loss $\mathcal{L}$ is given by:
\begin{equation}
\mathcal{L} =  \mathcal{L}_{rgb} + \lambda_{d} \cdot \mathcal{L}_{d} + \lambda_{n} \cdot \mathcal{L}_{n} + \mathcal{L}_{mv},
\end{equation}

\noindent where $ \lambda_{d} $ and $ \lambda_{n} $ control the relative contributions of depth and normal supervision, respectively.

\section{Experiments}

\begin{figure*}[t]
  \centering
   \includegraphics[width=\linewidth]{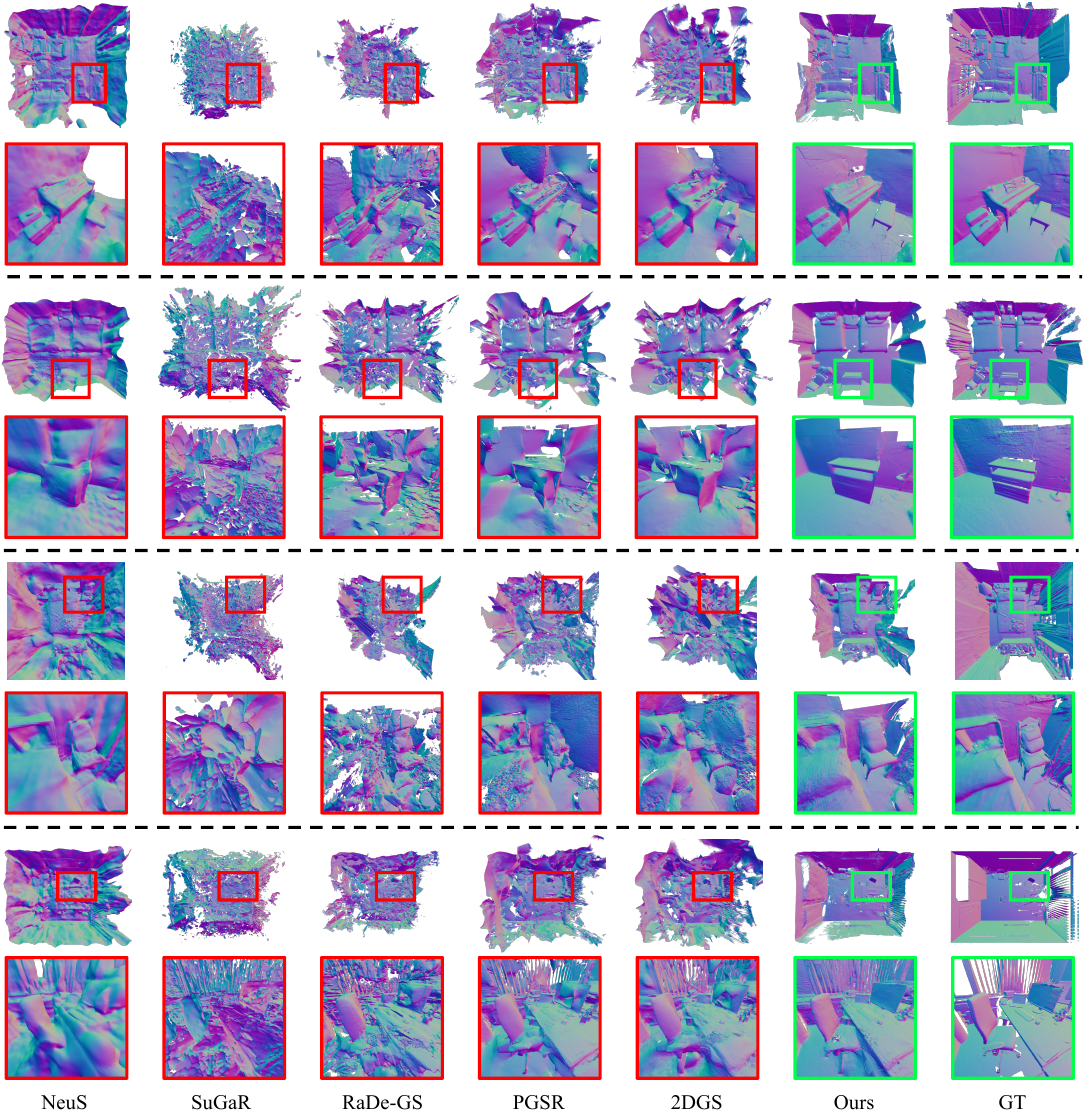}
   \caption{\textbf{Qualitative reconstruction comparisons.} For each indoor scene, the first row is the top view of the whole room, and the second row is the details of the masked region. }
   \label{fig:results_comparison}
\end{figure*}

\subsection{Experimental Setup}

\textbf{Dataset.} We evaluate the performance of our approach on reconstruction quality across 12 real-world indoor scenes from publicly available datasets: 8 scenes from ScanNet(V2) \cite{scannet} and 4 scenes from ScanNet++ \cite{scannet++}.

\noindent \textbf{Implementation Details.} Our training strategy and hyperparameters are consistent with the baseline 2DGS method to ensure comparability. We set $k = 10$, $\lambda_1 = 0.01$, $\lambda_{\cos} = 0.01$, $\lambda_{grad} = 0.5$,  $\lambda_{geo} = 0.05$, $\lambda_{pho} = 0.2$, $\lambda_{d} = 1.0$, $\lambda_{n} = 1.0$, in all our experiments. We render depth maps for all training views and then adopt TSDF fusion \cite{tsdf} for mesh extraction. We train all models for 30k iterations. All experiments are conducted on an NVIDIA RTX 4090 GPU to ensure consistent processing.

\noindent \textbf{Metrics.} Consistent with existing methods \cite{wang2022neuris, yu2022monosdf}, five standard metrics are employed to evaluate the quality of reconstructed meshes: Accuracy, Completion, Precision, Recall, and F-score.

\noindent \textbf{Baselines.} We compare our approach with several state-of-the-art methods, covering both neural volume rendering and Gaussian splatting techniques. The baselines include (1) Neural volume rendering methods: NeuS \cite{wang2021neus} and NeuralAngelo \cite{li2023neuralangelo}; (2) Gaussian splatting methods: 3DGS \cite{kerbl20233dgs}, SuGaR \cite{guedon2023sugar}, RaDe-GS \cite{zhang2024rade}, PGSR \cite{chen2024pgsr}, and 2DGS \cite{huang20242d}.

\subsection{Results Analysis}

\textbf{Qualitative Results.} To show the visualized reconstruction results of our method, we compare our 2DGS-Room with different reconstruction methods, including NeuS \cite{wang2021neus}, SuGaR \cite{guedon2023sugar}, RaDe-GS \cite{zhang2024rade}, PGSR \cite{chen2024pgsr}, 2DGS \cite{huang20242d}, and the ground truth. As illustrated in Figure \ref{fig:results_comparison}, our method exhibits significantly clearer scene structures, which is largely attributed to the seed-guided strategy. Additionally, thanks to the incorporation of depth and normal priors, the overall quality of our reconstructions is noticeably higher. In comparison with Gaussian-based methods, our method obtains a more visually coherent and accurate representation of the indoor scenes, with well-defined surfaces and consistent details across different views.

\noindent \textbf{Quantitative Results.} Quantitative results are presented in Table \ref{tab:2D_compare_quality}, showing a comprehensive comparison in geometry metrics on indoor scene datasets. On the ScanNet dataset, our method achieves the best results in all metrics. Compared to NeRF-based methods \cite{wang2021neus,li2023neuralangelo} which typically require over 20 hours to train a scene, our method significantly reduces training time, being approximately 30 times faster.

Since our method directly uses 2D Gaussians to represent scene surfaces, allowing the Gaussian splat to better adhere to the surface geometry, it outperforms 3DGS-based methods \cite{kerbl20233dgs,guedon2023sugar}. Furthermore, while 2DGS \cite{huang20242d} and some other methods \cite{chen2024pgsr,zhang2024rade} that employ depth strategies do improve geometric reconstruction quality, they still struggle in indoor scenes due to the complexity of spatial structures and the prevalence of textureless regions. By integrating seed-guided strategies and geometric constraints, our method enhances the accuracy of scene structure capture and achieves higher reconstruction quality, resulting in superior metrics.

As shown in Fig.~\ref{fig:results_comparison}, some methods \cite{guedon2023sugar,wang2021neus} produce noisy reconstructions with scattered floaters, and fail to represent the actual surfaces accurately due to the lack of geometric constraints. However, they may cover more ground truth data and thus achieve higher Accuracy than 2DGS on the ScanNet++ dataset in Table \ref{tab:2D_compare_quality}. Our method improves the structural coherence of the reconstruction, leading to a more accurate representation of the scene and a significant improvement in the Accuracy metric compared to 2DGS.


\subsection{Ablation Studies}

\begin{figure}[hb]
  \centering
   \includegraphics[width=\linewidth]{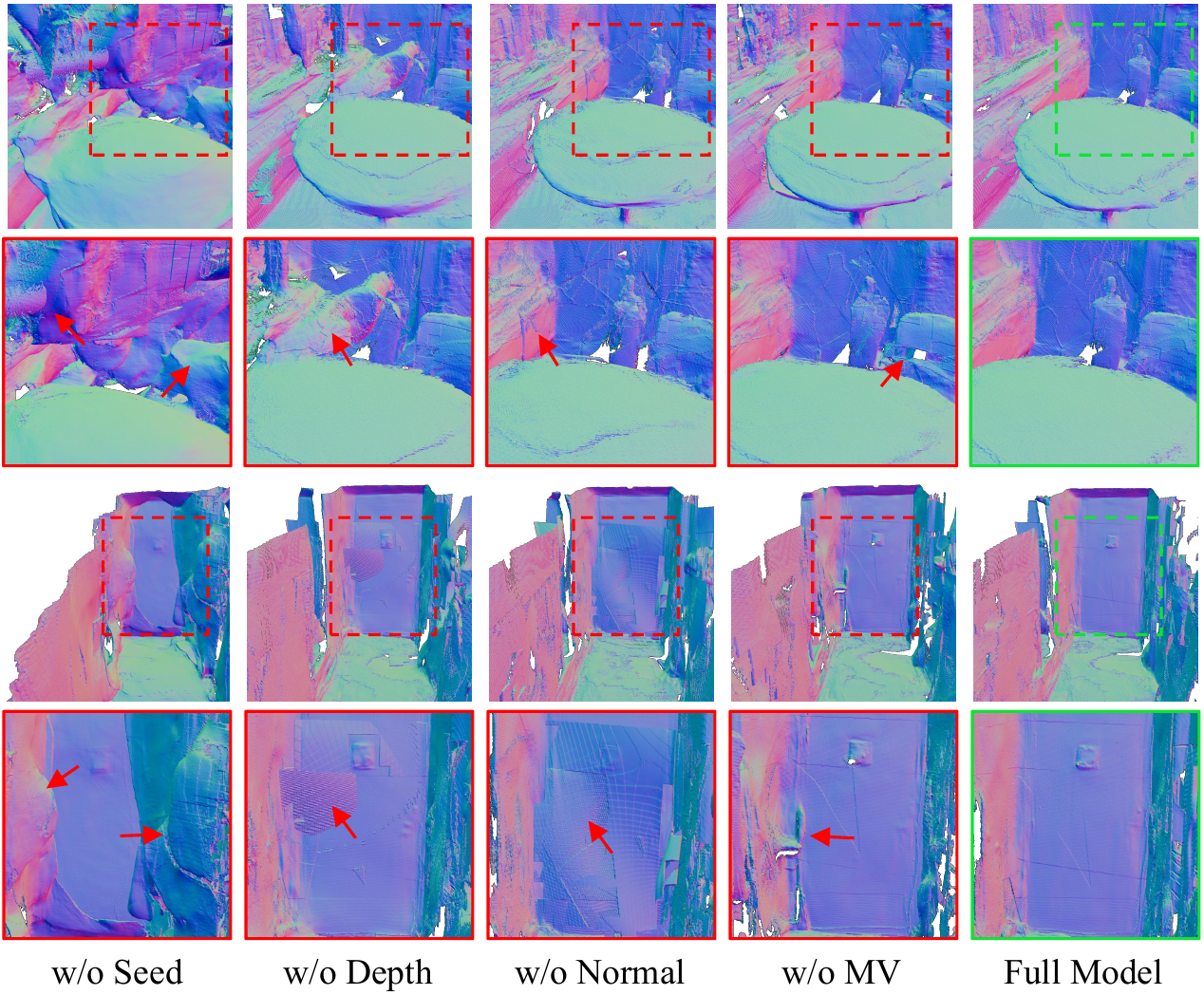}
   \caption{\textbf{Qualitative results of ablation study.} }
   \label{fig:ablation_details}
   \vspace{-0.4cm}
\end{figure}

To assess the individual contributions of each component in our model, we perform ablation studies on the ScanNet dataset. The quantitative results are reported in Table \ref{tab:ab_scannet} and Figure \ref{fig:ablation_details} shows the qualitative results. These allow us to isolate the impact of key elements on the overall reconstruction quality.

\begin{table}[ht]
    \centering
    \scalebox{0.9}
    {
    \begin{tabular}
    {lcccc>{\columncolor[gray]{0.902}}c}
    \toprule
    Method & Acc.↓ & Comp.↓ & Prec.↑ & Recall↑ & F-score↑ \\
    \midrule
    w/o Seed & 0.128 & 0.152 & 0.336 & 0.284 & 0.307 \\
    w/o Depth & 0.084 & 0.139 & 0.510 & 0.386 & 0.438 \\
    w/o Normal & 0.066 & 0.102 & 0.596 & 0.463 & 0.520 \\
    w/o MV & 0.055 & 0.092 & 0.644 & 0.508 & 0.566 \\
    Full model & \textbf{0.055} & \textbf{0.092} & \textbf{0.648} & \textbf{0.518} & \textbf{0.575} \\
    
    \bottomrule
    \end{tabular}
    }
    \caption{Results of the ablation study on ScanNet dataset. The best results are marked in \textbf{bold}.}
    \label{tab:ab_scannet}
    \vspace{-0.4cm}
\end{table}

\noindent \textbf{Seed Points Guidance.} Figure \ref{fig:ablation_details} shows that without seed points guidance, the scene lacks clear structural organization, leading to a significantly inflated and disorganized reconstruction. Adding this module enables our method to better capture the underlying geometric framework of indoor scenes, improving the F-score by 87.3\% in Table \ref{tab:ab_scannet}.


\noindent \textbf{Monocular Depth Supervision.} As shown in Figure \ref{fig:ablation_details}, removing depth supervision leads to spatial misalignments and unrealistic arrangements. Incorporating depth supervision significantly enhances geometric accuracy, achieving a 31.3\% F-score increase as reported in Table \ref{tab:ab_scannet}.


\noindent \textbf{Monocular Normal Supervision.} Removing normal supervision results in surface inconsistencies as shown in Figure \ref{fig:ablation_details}, with certain planar areas like walls, floors, and doors misaligned. Adding this module improves surface alignment, increasing the F-score by 10.6\% in Table \ref{tab:ab_scannet}.


\noindent \textbf{Multi-View Consistency Constraints.} Figure \ref{fig:ablation_details} reveals some Gaussians fail to align with the correct areas with the absence of multi-view constraints. Introducing this component reduces view-dependent inconsistencies to a certain degree, further enhancing the reconstruction quality.


\section{Conclusion}

We propose 2DGS-Room, a novel method for indoor scene reconstruction based on 2D Gaussian splatting by incorporating structural information from the scene to generate seed points, which guide the local Gaussian distributions. By leveraging geometric priors, we enhance the reconstruction quality of textureless regions and fine details in complex indoor environments. We also utilize multi-view consistency to reduce view-dependent inconsistencies to a certain degree. Extensive experiments show our method achieves superior performance compared with existing methods on multiple metrics and various indoor scenes.

{
    \small
    \bibliographystyle{ieeenat_fullname}
    \bibliography{main}
}

\clearpage
\setcounter{page}{1}
\maketitlesupplementary
\appendix

In this supplementary material, we provide the following components:
\begin{itemize}
    \item Definitions of the 3D geometry metrics used to evaluate reconstruction quality in Sec.~\ref{sec: metrics}.
    \item Additional details of the datasets, training configuration, and the iteration schedule for key modules in Sec.~\ref{sec: implementation details}. 
    \item Additional qualitative results, including mesh comparison, ablation results, and rendering comparison in Sec.~\ref{sec: comparison}.
\end{itemize}


\section{Definitions of Eevaluation Metrics} \label{sec: metrics}

We evaluate our method using five widely-used 3D geometry metrics: Accuracy, Completion, Precision, Recall, and F-score, defined in Table \ref{tab:metric_defs}. These metrics collectively assess the geometric fidelity of the reconstructed point clouds by measuring the alignment between the predicted and ground truth point clouds. 

Accuracy measures the average distance between reconstructed points and the ground truth, with smaller values indicating better alignment. Completion assesses how well the reconstruction covers the ground truth, where lower values are better. Precision and Recall evaluate the proportion of points within a set threshold, with higher values indicating better performance. F-score, the harmonic mean of Precision and Recall, provides a balanced measure of reconstruction quality, where higher values reflect superior results.

\begin{table}[htb]
\centering
\begin{tabular}{ll}
\toprule[1pt]
Metric & Definition \\
\midrule
Acc. & $\mbox{mean}_{c \in C}(\min_{c^*\in C^*}||c-c^*||)$ \\
Comp. & $\mbox{mean}_{c^* \in C^*}(\min_{c\in C}||c-c^*||)$ \\
Prec. & $\mbox{mean}_{c \in C}(\min_{c^*\in C^*}||c-c^*||<.05)$ \\
Recall & $\mbox{mean}_{c^* \in C^*}(\min_{c\in C}||c-c^*||<.05)$ \\
zoF-score & $\frac{ 2 \times \text{Prec} \times \text{Recall} }{\text{Prec} + \text{Recall}}$ \\
\bottomrule[1pt]
\end{tabular}
\caption{\textbf{Definitions of 3D metrics.} $c$ and $c^*$ are the predicted and ground truth point clouds.}
\label{tab:metric_defs}
\vspace{-0.4cm}
\end{table}

\section{Additional Implementation Details} \label{sec: implementation details}

\textbf{Datasets.} As described in the main paper, the quantitative evaluation metrics are derived from results tested two datasets. Specifically, we select 8 scenes from the ScanNet dataset \cite{scannet}: scene0050\_00, scene0085\_00, scene0114\_02, scene0580\_00, scene0603\_00, scene0616\_00, scene0617\_00, scene0721\_00, and 4 scenes from the ScanNet++ dataset \cite{scannet++}: 8b5caf3398, 8d563fc2cc, 41b00feddb, b20a261fdf.

\noindent \textbf{Training details.} For all scenes, our seed-guided optimization is performed between 1,500 and 15,000 iterations. We set $N_g=100$ for the gradient-guided growth and $N_\alpha=100$ for the pruning strategy. Depth supervision and normal supervision are applied consistently from the first iteration through to the end of training, providing continuous geometric constraints. The multi-view consistency constraint is introduced after 7,000 iterations, once the foundational structure has been established, to further improve view alignment.


\section{Additional Qualitative Results} \label{sec: comparison}

\subsection{Additional Ablation Results}

\begin{figure}[b]
  \centering
   \includegraphics[width=\linewidth]{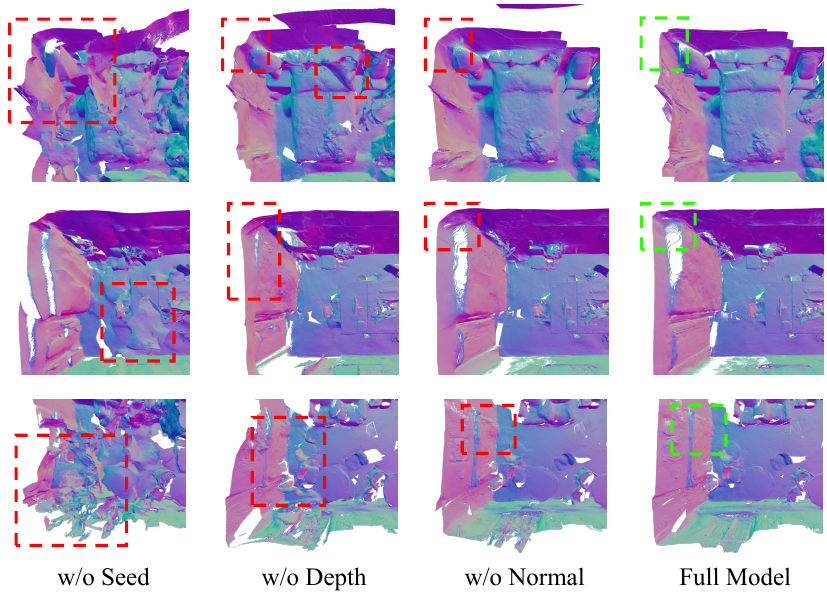}
   \caption{\textbf{Additional qualitative results of ablation study.}
   }
   \label{fig:x_ablation}
\end{figure}

To complement the local detail comparisons in the main paper, we provide additional ablation results focusing on the overall scene structure in Figure \ref{fig:x_ablation}. These visualizations highlight the contributions of key components, including the seed points guidance, monocular depth supervision, and monocular normal supervision. The multi-view consistency constraints are primarily designed to further mitigate floating artifacts in certain scenarios, which have a limited impact on the overall structure. Therefore, they are not included in these structural comparisons. Their effectiveness is instead reflected in the qualitative results shown in Figure \ref{fig:ablation_details} and the quantitative metrics presented in Table \ref{tab:ab_scannet} of the main paper.

When the seed points guidance strategy is removed, the reconstructed objects appear fused together, with unclear boundaries, compromising the scene’s structural clarity. Without depth supervision, objects exhibit depth misalignments, leading to unrealistic spatial arrangements. Similarly, excluding normal supervision results in uneven surfaces, especially on planar regions like walls, where visible curvature or misalignment artifacts occur.

\subsection{Additional Qualitative Comparison}

\begin{figure*}[hb]
  \centering
   \includegraphics[width=\linewidth]{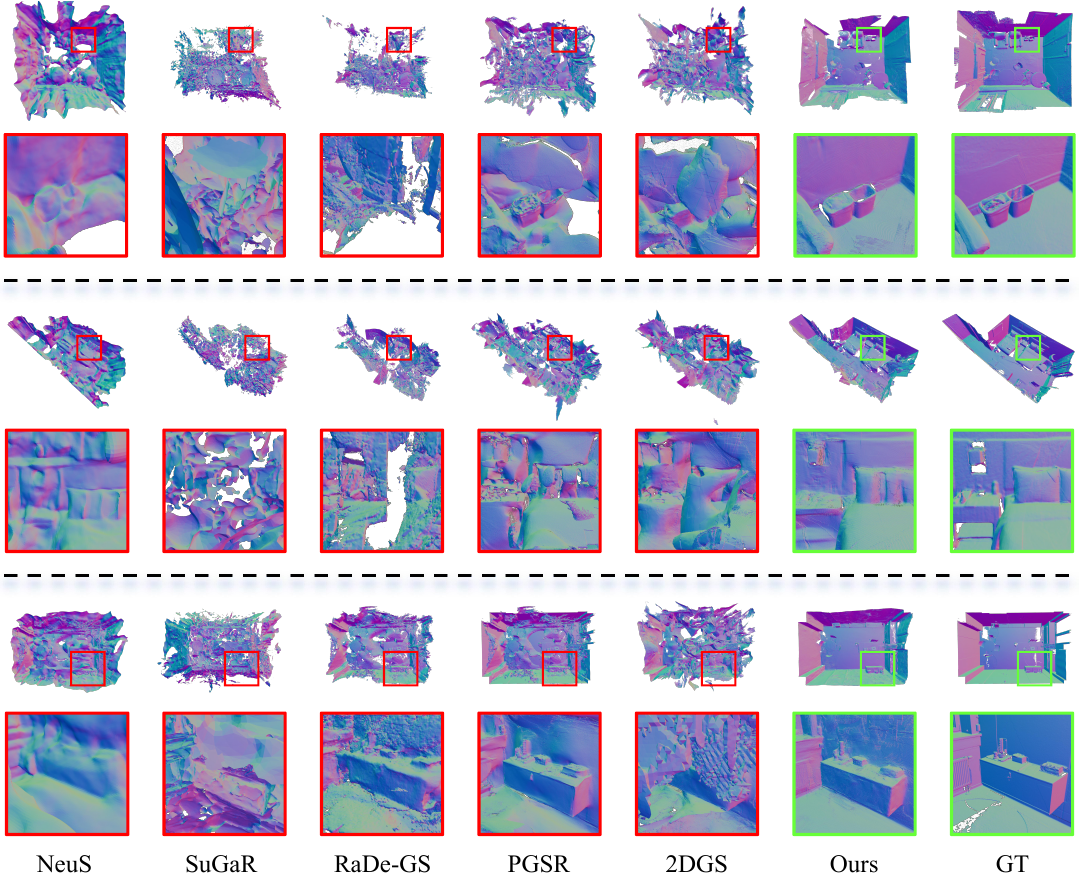}
   \caption{\textbf{Additional qualitative reconstruction comparison.} For each indoor scene, the first row is the top view of the whole room and the second row is the details of the masked region.}
   \label{fig:x_comparison}
\end{figure*}

In addition to the four indoor scenes shown in the main paper, we further include qualitative reconstruction comparison results of the different methods \cite{wang2021neus,guedon2023sugar,zhang2024rade,chen2024pgsr,huang20242d} on additional scenes from ScanNet and ScanNet++. As demonstrated in Figure \ref{fig:x_comparison}, our method significantly outperforms other approaches in capturing global structures, preserving fine-grained details as well as reducing artifacts in textureless regions.

\subsection{Rendering Comparison}

\begin{figure*}[hb]
  \centering
   \includegraphics[width=\linewidth]{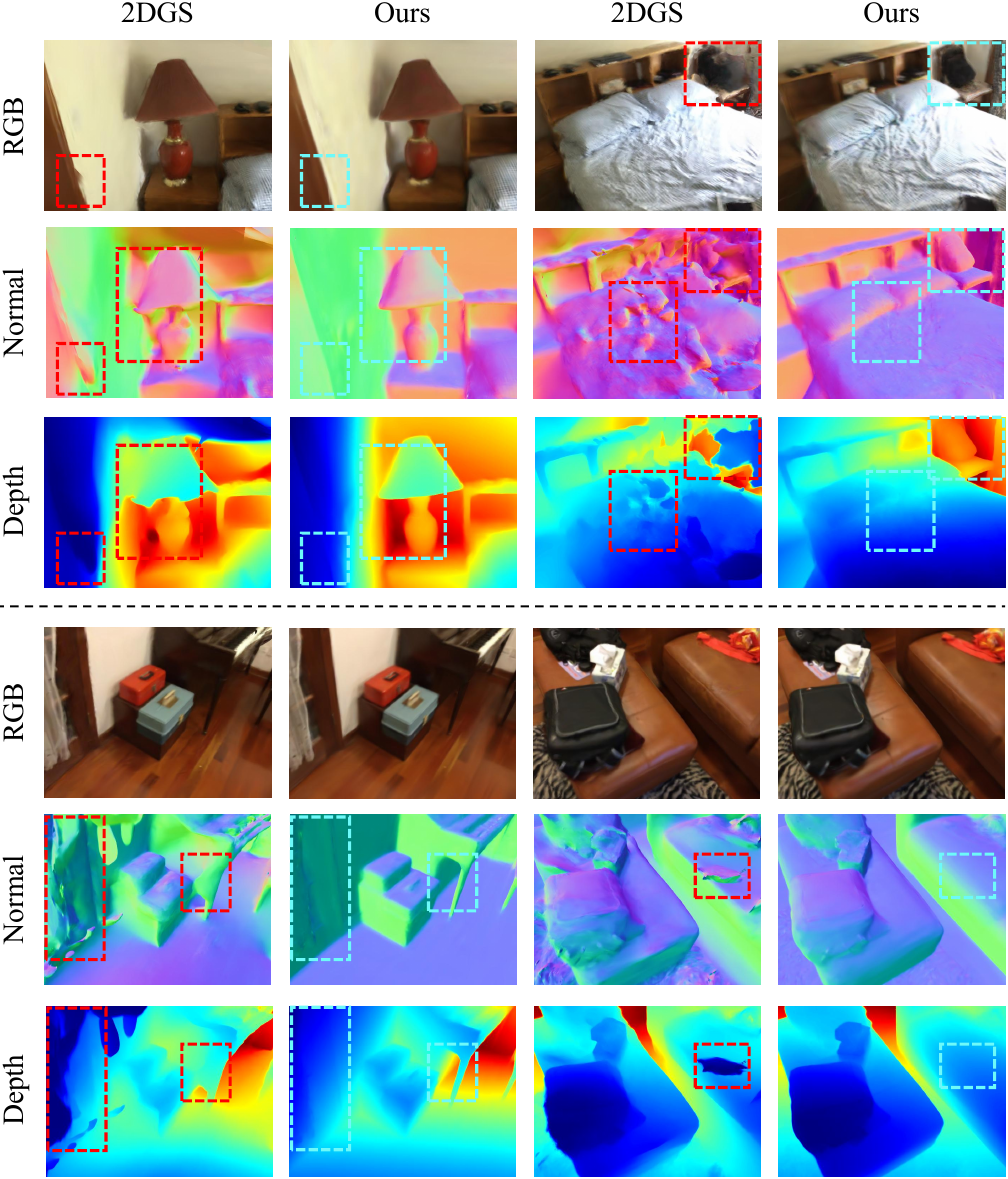}
   \caption{\textbf{Rendering comparison on the ScanNet dataset (scene0580 and scene0050).}}
   \label{fig:x_render_1}
\end{figure*}

\begin{figure*}[hb]
  \centering
   \includegraphics[width=\linewidth]{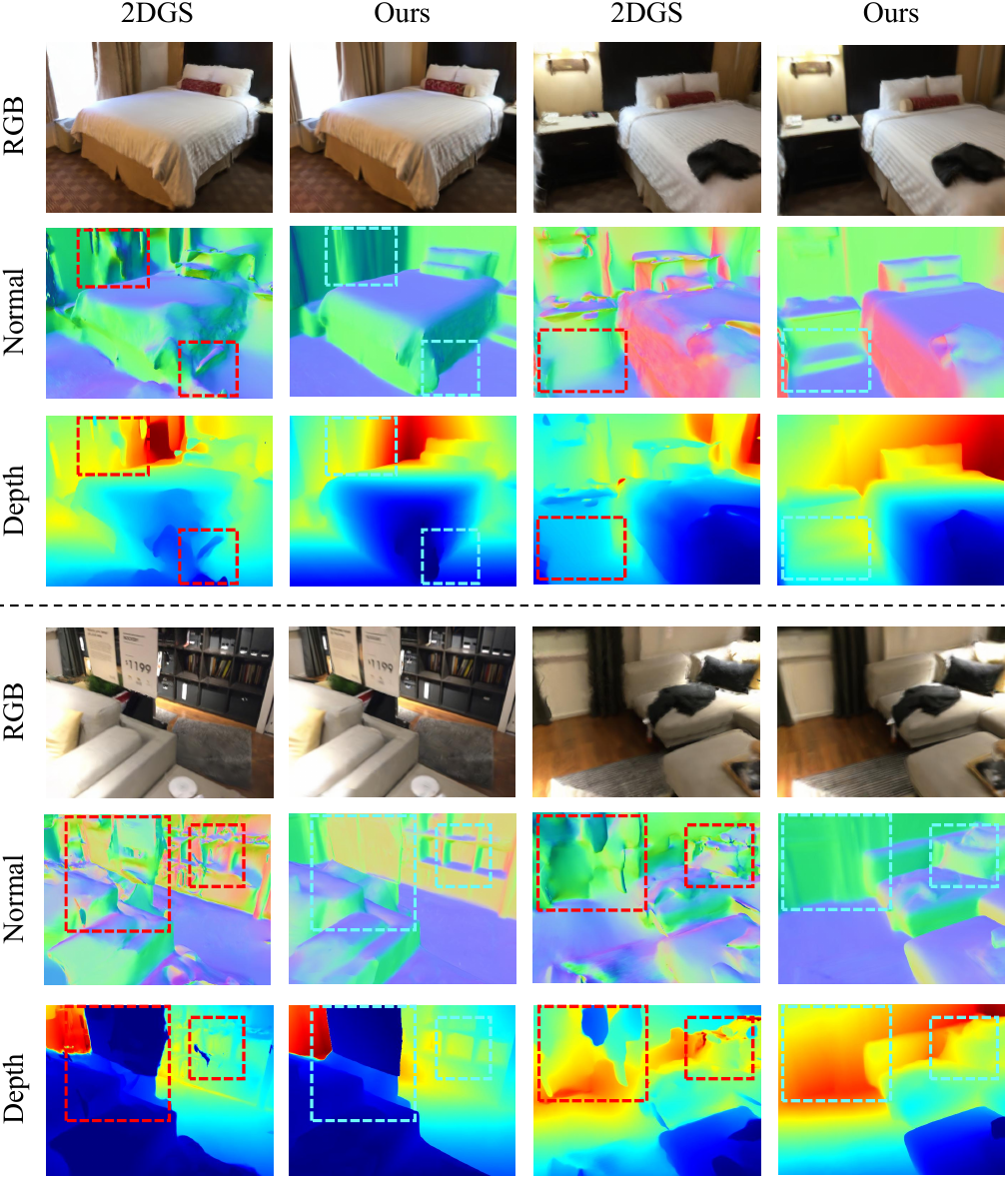}
   \caption{\textbf{Rendering comparison on the ScanNet dataset (scene0085 and scene0617).} }
   \label{fig:x_render_2}
\end{figure*}

\begin{figure*}[hb]
  \centering
   \includegraphics[width=\linewidth]{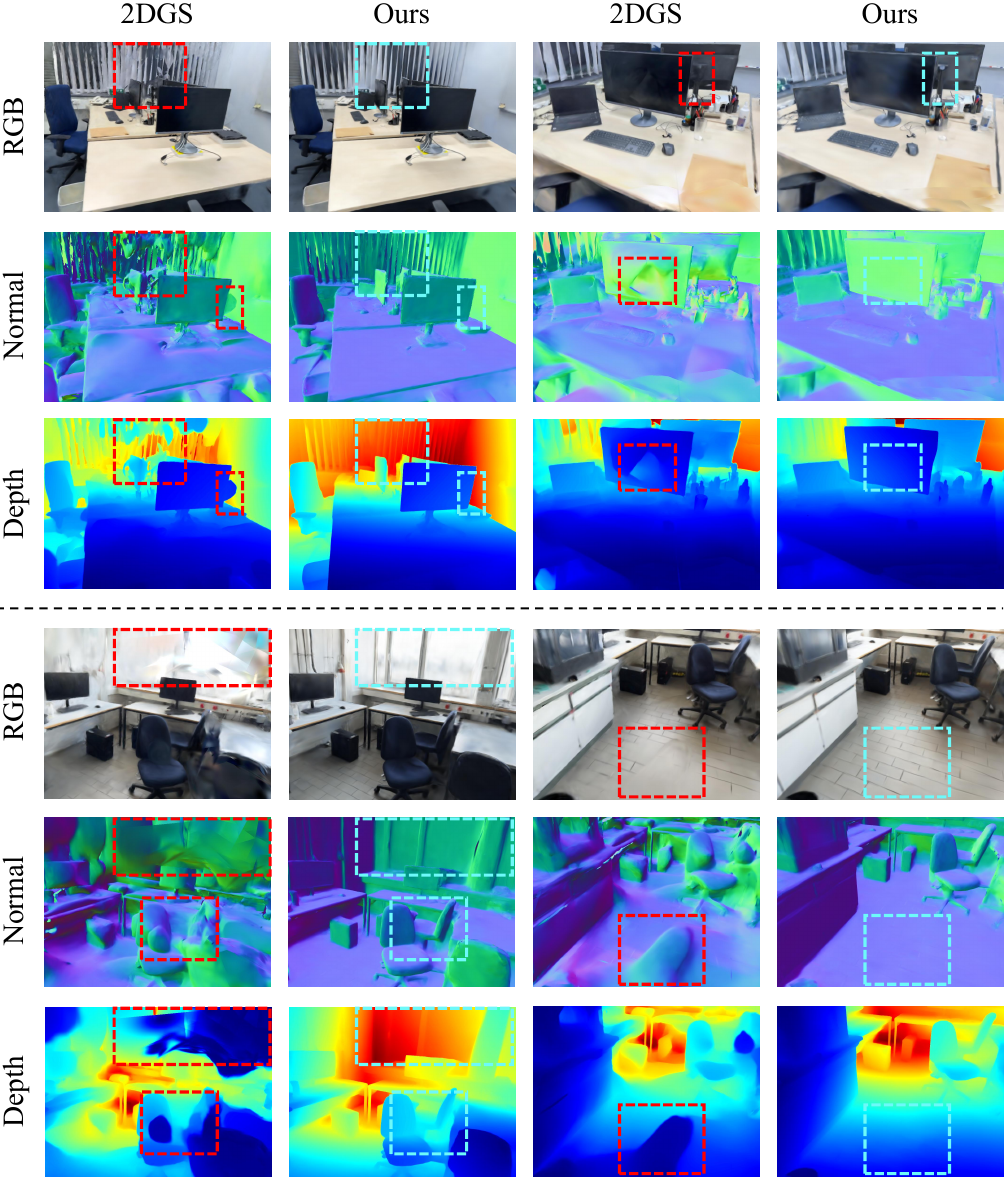}
   \caption{\textbf{Rendering comparison on the ScanNet++ dataset (8d563fc2cc and 41b00feddb).} }
   \label{fig:x_render_3}
\end{figure*}

We also provide extensive rendering results comparing our 2DGS-Room with 2DGS across various scenes and viewpoints from the ScanNet and ScanNet++ datasets in Figures \ref{fig:x_render_1}, \ref{fig:x_render_2}, and \ref{fig:x_render_3}. Rendered RGB, depth, and normal maps are shown for visual comparison. Our method achieves significant improvements in the rendering quality of depth and normal maps, showcasing smoother transitions and more accurate surface details. Furthermore, the quality of the RGB images rendered by our method remains robust and shows clear advantages over 2DGS in challenging scenarios, such as handling fine details and varying lighting conditions. This demonstrates the effectiveness of our method in achieving superior geometric reconstructions while maintaining photometric accuracy.

\end{document}